\documentclass[journal]{IEEEtran}
\usepackage{amsmath,amsfonts}
\usepackage{amsthm,amsmath,amssymb}
\usepackage{mathrsfs}
\usepackage{algorithmic}
\usepackage{algorithm}
\usepackage{array}
\usepackage[caption=false,font=normalsize,labelfont=sf,textfont=sf]{subfig}
\usepackage{tabularx}
\usepackage{textcomp}
\usepackage{stfloats}
\usepackage{multirow}
\usepackage{multicol}
\usepackage{url}
\usepackage{verbatim}
\usepackage{graphicx}
\usepackage{cite}
\usepackage{booktabs} 
\usepackage{hyperref}
\usepackage{orcidlink}
\begin{document}

\title{Unpaired Object-Level SAR-to-Optical Image Translation for Aircraft with Keypoints-Guided Diffusion Models}
\author{Ruixi You, ~\IEEEmembership{Graduate Student Member,~IEEE}, Hecheng Jia, ~\IEEEmembership{Member,~IEEE}, Feng Xu\orcidlink{0000-0002-7015-1467}, ~\IEEEmembership{Senior Member,~IEEE}

\thanks{ This work was supported by xxx. \textit{(Corresponding author: Feng Xu.)} 

The authors are with the Key Laboratory for Information Science of Electromagnetic Waves (Ministry of Education), School of Information Science and Technology, Fudan University, Shanghai 200433, China. (fengxu@fudan.edu.cn)}}

\markboth{Journal of \LaTeX\ Class Files,~Vol.~18, No.~9, February~2025}%
{How to Use the IEEEtran \LaTeX \ Templates}

\maketitle

\begin{abstract}
Synthetic Aperture Radar (SAR) imagery provides all-weather, all-day, and high-resolution imaging capabilities but its unique imaging mechanism makes interpretation heavily reliant on expert knowledge, limiting interpretability, especially in complex target tasks. Translating SAR images into optical images is a promising solution to enhance interpretation and support downstream tasks. Most existing research focuses on scene-level translation, with limited work on object-level translation due to the scarcity of paired data and the challenge of accurately preserving contour and texture details.
To address these issues, this study proposes a keypoint-guided diffusion model (KeypointDiff) for SAR-to-optical image translation of unpaired aircraft targets. This framework introduces supervision on target class and azimuth angle via keypoints, along with a training strategy for unpaired data. Based on the classifier-free guidance diffusion architecture, a class-angle guidance module (CAGM) is designed to integrate class and angle information into the diffusion generation process. Furthermore, adversarial loss and consistency loss are employed to improve image fidelity and detail quality, tailored for aircraft targets. During sampling, aided by a pre-trained keypoint detector, the model eliminates the requirement for manually labeled class and azimuth information, enabling automated SAR-to-optical translation.
Experimental results demonstrate that the proposed method outperforms existing approaches across multiple metrics, providing an efficient and effective solution for object-level SAR-to-optical translation and downstream tasks. Moreover, the method exhibits strong zero-shot generalization to untrained aircraft types with the assistance of the keypoint detector.
\end{abstract}


\section{Introduction}
\IEEEPARstart{S}{ynthetic}  Aperture Radar (SAR) imagery, with its all-weather, all-time, and high-resolution imaging capabilities, has been widely used in target detection and monitoring tasks in complex environments\cite{SAR_tutorial}. Compared to optical sensors, SAR remains operational under conditions such as cloud cover or insufficient light, making it an indispensable tool for monitoring in challenging environments. However, the unique microwave-based imaging mechanism of SAR results in images that are characterized by inherent speckle noise and lack the clear contours and texture details typically seen in optical images. This results in poor image readability and significantly hampers automatic interpretation, which heavily relies on expert knowledge\cite{MicrowaveVision_review,Gestalt}. The challenge is especially pronounced in interpretation tasks for complex targets such as aircraft and ships, where SAR images often present as discrete scattering clusters instead of clear, closed contours with colorful texture information, making it difficult to interpret fine details. As a result, SAR images are less intuitive for human understanding, severely affecting both the efficiency and accuracy of interpretation tasks, particularly when dealing with small or intricate targets such as aircraft.

Translating SAR images into optical images presents a powerful solution for enhancing the interpretability and visual appeal of SAR data, making it more aligned with human visual perception. Unlike SAR images, which often suffer from low readability due to inherent speckle noise, lack of clear texture, and complex background interference, optical images are more intuitive and familiar for human observers. By converting SAR images into optical representations, this translation significantly improves both the visual quality and accessibility of SAR data, making it easier to interpret and analyze. This enhancement is crucial for downstream tasks such as target detection, classification, and monitoring, where accurate and efficient data interpretation is vital. As a result, SAR-to-optical image translation has gained increasing attention in recent years as a key area of research, as it holds the potential to bridge the gap between the technical advantages of SAR imaging and the more accessible nature of optical imagery, thereby opening up new possibilities for remote sensing applications in complex environments

However, translating SAR images into optical images is a complex cross-modal generation problem. The main challenge arises from the substantial differences between the imaging principles of SAR and optical systems. These differences lead to distinct disparities in the resulting image characteristics, which complicate the translation task. SAR images are more prone to geometric distortions due to the unique radar imaging process and motion effects, while optical images can achieve high geometric fidelity through well-established techniques. Additionally, SAR images typically lack the detailed textures found in optical images, as they primarily capture information based on backscattering rather than surface features. Furthermore, the nonlinear brightness distribution in SAR images, caused by varying scattering properties, is another challenge, as optical images follow a more predictable light-to-color transformation. These inherent differences in image characteristics, including the lack of smooth gradients, textures, and perceptual cues, make the translation from SAR to optical images particularly difficult.

In recent years, deep generative models have garnered significant attention for cross-modal image generation tasks. From the early development of Generative Adversarial Networks (GANs)\cite{GAN} to the more recent advancements in diffusion models, generative techniques have rapidly evolved, significantly advancing the field of SAR-to-optical image translation\cite{review_RS,review_SAR,SAR_GENAI_review}. GANs, through adversarial training, enable efficient image generation but face challenges such as training instability and mode collapse. In contrast, diffusion models\cite{diffusion_ddpm} have emerged as a powerful alternative, generating high-quality images by progressively adding and removing noise, offering notable advantages in terms of diversity and stability in the generated images. Moreover, diffusion models are highly versatile in incorporating conditional information, such as class, semantic, or spatial features, making them an attractive solution for SAR-to-optical translation tasks.

\begin{figure}[tb]
  \centering
\includegraphics[width=\linewidth]{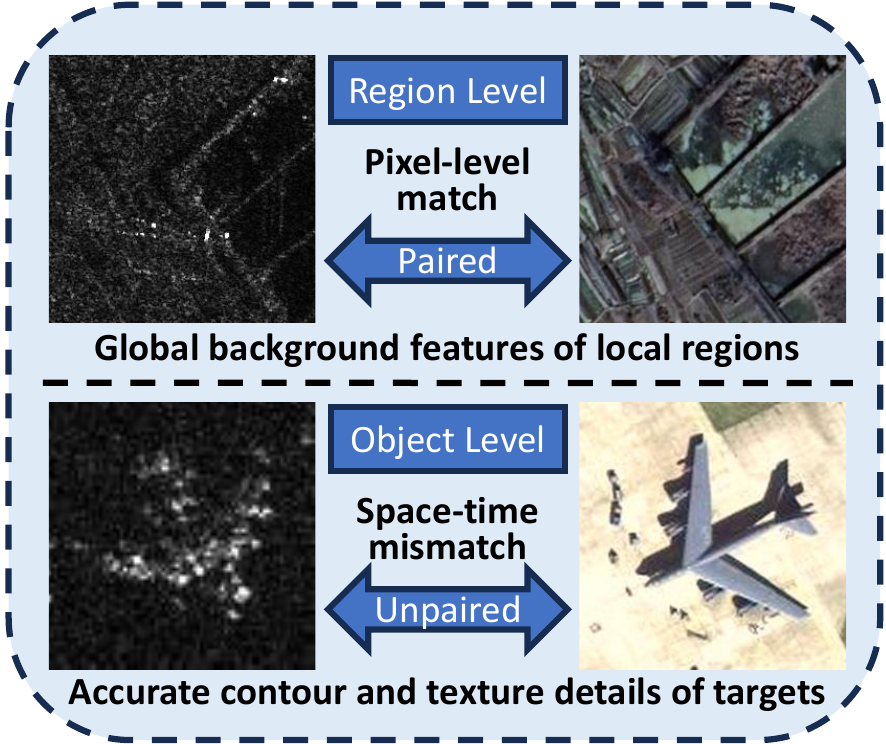}
  \caption{Illustrating the comparison between region-level and object-level SAR-to-optical image translation. Object-level translation faces challenges including unpaired training data and the demand for accurate recovery of contours and texture details.}
  \label{fig:object-level}
\end{figure}

While considerable progress has been made in scene-level SAR-to-optical translation~\cite{gan_Reyes,gan_cran,GAN_Atrous,GAN_Fine-Grained,GAN_Edge-Preserving,GAN_Hierarchical-Latent-Features,GAN_Wavelet-Feature,GAN_vit,GAN_improved-CGAN,GAN_Texture-and-Structure,GAN_Coarse-to-Fine,gan_semi,gan_cloud,diffusion_bai-condition,diffusion_bai-color,diffusion_Color-Memory,diffusion_language,diffusion_Spatial-Frequency,diffusion_bai-distillation} , the task of object-level translation remains relatively underexplored. Generally, object-level translation for aircraft introduces two primary challenges as shown in Fig.~\ref{fig:object-level}:\begin{itemize}
\item[1)] \textit{Lack of spatio-temporally aligned data}. Scene-level tasks often rely on broader environmental features, where paired data is more readily available, and temporal alignment between SAR and optical images is less stringent. In contrast, object-level translation demands a much higher level of precision, as it requires matching specific target information across both modalities. The lack of spatio-temporal alignment in paired data at the object level hinders the ability to learn robust mappings effectively. 
\item[2)] \textit{Requirement for fine-grained category-specific features}. Unlike scene-level tasks that focus on holistic representations, object-level translation necessitates accurate reconstruction of detailed target attributes, such as contours and textures, which are intricately linked to the object’s category. As a result, achieving high fidelity and accurate feature reconstruction for complex targets remains a significant challenge. These factors make SAR-to-optical object-level translation a critical yet under-researched area.
\end{itemize}

To address the aforementioned challenges, this paper proposes a keypoint-supervised diffusion model framework (KeypointDiff) for SAR-to-optical translation of unpaired aircraft targets. The proposed approach is based on two key insights to tackle the aforementioned issues: first, to alleviate the data mismatch problem, a training strategy is designed that leverages the unique viewpoint of SAR images and the rotation invariance of optical images\cite{Gestalt}; second, to address the difficulty in recovering target contours and textures, specialized modules and loss functions are developed to  enhance the model's performance on these inherent features.

Specifically, first, a keypoint-supervised training strategy is introduced to handle unpaired data effectively without requiring paired datasets, which are often difficult to obtain. This strategy capitalizes on the unique characteristics of SAR images and optical images, helping the model to handle unpaired data effectively, enabling robust learning from unpaired samples. Building on this, the Class-Angle Guidance Module (CAGM) is integrated into the classifier-free diffusion model architecture, directly encoding target class and azimuth angle information into the generation process. This improves the model's ability to generate more accurate and realistic optical images. Additionally, specialized loss functions are designed to further address the inherent challenges of SAR-to-optical translation, particularly in preserving high-fidelity contours and textures specific to aircraft targets. Finally, during the sampling phase, a pre-trained keypoint detector is utilized to eliminate the need for target class and azimuth angle labels, enabling fully automated translation and streamlining the inference process.

In summary, the main contributions of this work are as follows:\begin{itemize}
\item[1)] For unpaired object-level SAR-to-optical translation, a conditional diffusion model framework specifically designed for aircraft targets, named KeypointDiff, is proposed. Experimental results demonstrate that the proposed  method outperforms existing approaches across multiple evaluation metrics, generating high-fidelity optical aircraft images.
\item[2)] To mitigate the issue of unpaired data, a keypoint-based training strategy is introduced, eliminating the need for paired datasets.
\item[3)] For the accurate restoration of aircraft target contours and textures, specialized modules named CAGM are developed to integrate aircraft class and azimuth angle to the generation process and the joint loss functions tailored to the unique characteristics of aircraft targets.
\item[4)] During sampling, a pre-trained keypoint detector is leveraged to enable fully automated translation without the need for target class or azimuth angle labels, meanwhile exhibiting strong zero-shot generalization capabilities, enabling the model to handle untrained aircraft target types effectively. 
\end{itemize}

The rest of this paper is arranged as: Section \ref{sec:related} revisits the prior studies concerning sar-to-optical image translation and diffusion models. Section \ref{sec:method} elaborates on details of the proposed method KeypointDiff.  Section \ref{sec:experiment} presents experimental results and a thorough analysis. Section \ref{sec:conclusion} concludes the paper.

\section{Related Works}
\label{sec:related}

\subsection{SAR-to-OPT Image Translation}
Synthetic Aperture Radar (SAR) to optical (OPT) image translation has garnered increasing attention in recent years due to its significant potential in remote sensing applications. Traditional approaches treat this task as a colorization problem, relying on modeling the scattering properties of individual pixels in SAR images to enhance their visual appearance~\cite{color_scatter, color_xu}. However, such rule-based colorization methods, akin to classification tasks, struggle to achieve true optical image generation, leading to suboptimal visual quality and low accuracy.

Currently, most SAR-to-OPT image translation studies focus on deep learning methods based on Generative Adversarial Networks (GANs). GANs utilize adversarial training to generate realistic target-domain images and have demonstrated excellent performance in cross-domain image translation tasks. Reyes \emph{et al.}~\cite{gan_Reyes} made an early attempt at implementing SAR-to-OPT image translation using CycleGAN. 
From the perspective of network structure optimization, Fu \emph{et al.}~\cite{gan_cran} proposed multiscale residual connections to improve SAR-to-optical translation accuracy. Turnes \emph{et al.}~\cite{GAN_Atrous} enhanced cGAN architectures using atrous convolutions and spatial pyramid pooling. Yang \emph{et al.}~\cite{GAN_Fine-Grained} introduced fine-grained strategies for better image detail reproduction. 
In feature utilization, Guo \emph{et al.}~\cite{GAN_Edge-Preserving} optimized generation using edge gradient information, while Wang \emph{et al.}~\cite{GAN_Hierarchical-Latent-Features} improved quality by aligning hierarchical latent features. Other works explored innovative approaches such as wavelet feature learning~\cite{GAN_Wavelet-Feature}, ViT-CNN feature fusion~\cite{GAN_vit}, and dual-generator networks for texture-structure fusion~\cite{GAN_Texture-and-Structure}.
For training strategy optimization, Lee \emph{et al.}~\cite{GAN_Coarse-to-Fine} adopted a two-stage generation strategy for coarse-to-fine optical image generation. Du \emph{et al.}~\cite{gan_semi} proposed a semi-supervised framework to integrate aligned and unaligned SAR-OPT pairs. Besides, Darbaghshahi \emph{et al.}~\cite{gan_cloud}successfully applied SAR-to-OPT translation to dehazing tasks with a dual-GAN network via dilated residual inception blocks.
These advancements collectively highlight diverse strategies for improving SAR-to-OPT translation, spanning network architecture, feature utilization, and training methodologies.

Recently, Diffusion Models have emerged as a generative framework for progressive denoising, showing promise in SAR-to-OPT image translation tasks. By modeling noise in a reverse process, diffusion models can generate high-quality images, achieving better results in SAR-to-OPT translation. Bai \emph{et al.}~\cite{diffusion_bai-condition} pioneered the use of conditional diffusion models for SAR-to-OPT translation, effectively mitigating common translation issues such as color shift~\cite{diffusion_bai-color}. Guo \emph{et al.}~\cite{diffusion_Color-Memory} further optimized color feature extraction modules to significantly enhance the quality of generated images. Do \emph{et al.}~\cite{diffusion_language} leveraged pre-trained text and image encoders to guide the training of diffusion models. Qin \emph{et al.}~\cite{diffusion_Spatial-Frequency} improved model structure and loss functions from the perspective of spatial-frequency refinement, achieving better results in translation tasks. Furthermore, to address the efficiency challenges of diffusion models, Bai \emph{et al.}~\cite{diffusion_bai-distillation} incorporated distillation techniques to optimize generation speed.

Despite the progress made in SAR-to-OPT translation, existing methods have rarely explored object-level translation. The primary challenges lie in the difficulty of recovering contours and textures in SAR and OPT target regions and the lack of paired datasets. To address these limitations, this study designs targeted training strategies and introduces supervision and optimization mechanisms for target regions, achieving unpaired object-level SAR-to-OPT image translation with enhanced fidelity and geometric consistency.

\subsection{Unpaired Image-to-Image Translation}
Image-to-image (I2I) translation is a significant research direction in computer vision, aiming to convert images from a source domain to a target domain. Classical methods, such as Pix2Pix~\cite{unpaired_pix2pix}, rely on paired training data and adopt an end-to-end supervised approach to achieve cross-domain mapping. However, in many practical scenarios, obtaining fully aligned and high-quality paired data is challenging or even infeasible, leading to the growing interest in unpaired I2I translation tasks.

Unpaired I2I translation approaches generally achieve cross-domain mapping by reducing the domain gap between the source and target domains. GAN-based methods represent one of the most prominent directions in this field. CycleGAN~\cite{unpaired_cyclegan} introduces a cycle consistency loss to ensure consistency between forward translation and backward mapping, effectively preserving bidirectional structural consistency. Building on this, MUNIT~\cite{unpaired_MUNIT} enables multimodal translation by disentangling content and style for more flexible mappings. DistanceGAN~\cite{unpaired_distanceGAN} incorporates mutual information regularization to reduce distributional differences between domains. GcGAN~\cite{unpaired_GcGAN}, on the other hand, focuses on geometric consistency, ensuring spatial alignment between domains. 
In addition to GAN-based methods, contrastive learning has been introduced into unpaired I2I tasks. CUT~\cite{unpaired_CUT} enhances translation quality by constructing positive and negative sample pairs, while NEGCUT~\cite{unpaired_NEGCUT} further optimizes the selection of negative samples to improve generation performance.

In recent years, diffusion models have garnered increasing attention in unpaired I2I translation tasks in many areas including medical imaging~\cite{unpaired_contourdiff}. EGSGE~\cite{unpaired_egsde} leverages energy guidance to optimize the stochastic differential equations (SDE) of diffusion models, significantly improving the quality and stability of the generated images. UNSB~\cite{unpaired_UNSB} adopts the Schrödinger Bridge (SB) framework to learn an SDE for efficient translation between cross-domain distributions.

For the unpaired I2I problem in remote sensing image tasks studied in this paper, the unique top-down perspective of remote sensing images provides a novel opportunity. By incorporating target keypoint supervision as geometric constraints, this article proposes a dynamic pairing strategy that enables unpaired training data to progressively align with the target domain distribution during optimization. This approach not only improves translation quality but also enhances geometric consistency and detail preservation in the target regions.

\begin{figure*}[tb]
  \centering
\includegraphics[width=\linewidth]{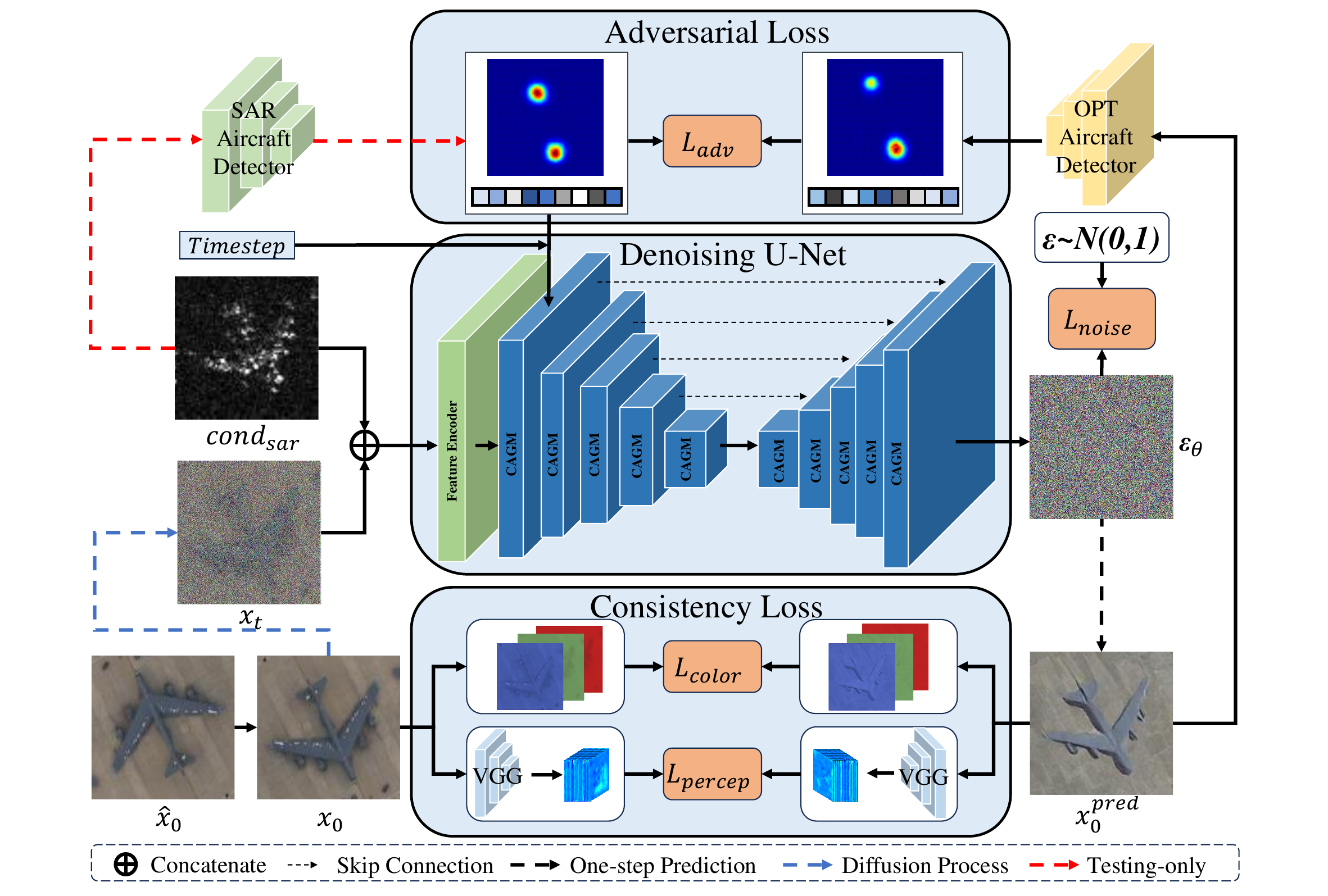}
\caption{Overview of the KeypointDiff framework, illustrating the core denoising U-Net architecture, the integration of customized loss functions , and the keypoints-based supervision strategy for both training and testing phases. }
  \label{fig:overview}
\end{figure*}

\subsection{Diffusion Models}

Diffusion models have demonstrated strong capabilities in various generation tasks, including image synthesis, text-to-image generation, and video generation~\cite{diffusion_review-acm, diffusion_review-tpami,diffusion_CLE}. Compared to traditional Generative Adversarial Networks (GANs), diffusion models employ a stepwise denoising process to generate samples, which enables them to stably capture data distributions and ensures a more robust training process, effectively avoiding common issues in GANs such as mode collapse.

The Denoising Diffusion Probabilistic Model (DDPM)~\cite{diffusion_ddpm} is a pioneering work in diffusion models that achieves high-quality image synthesis by modeling noise incrementally during the forward process and denoising it in the reverse process. Building upon DDPM, Improved DDPM~\cite{diffusion_improved-ddpm} introduced optimized training efficiency and sampling strategies, further enhancing the model's performance. Subsequently, Guided Diffusion~\cite{diffusion_guided-diffusion} incorporated conditional generation mechanisms by utilizing a pre-trained classifier to guide the generation process, enabling more controllable sample outputs. Classifier-Free Guidance (CFG)~\cite{diffusion_cfg} further advanced this framework by removing the dependency on pre-trained classifiers. Instead, CFG employs a linear combination of unconditional and conditional generation probabilities, improving both the flexibility and quality of the generated samples.

Significant optimizations have also been achieved in sampling strategies. Denoising Diffusion Implicit Models (DDIM)~\cite{diffusion_ddim} introduced non-Markovian sampling, simplifying the sampling process and improving efficiency. The respace strategy proposed in Improved DDPM~\cite{diffusion_improved-ddpm} optimized linear diffusion sampling by adjusting timesteps. Additionally, Diffusion Model Distillation~\cite{diffusion_distillation} applied distillation techniques to train a student model capable of generating high-quality images in a single step or with fewer steps, dramatically accelerating the inference speed of diffusion models.

Building upon the CFG structure, this work proposes a high-fidelity generation framework that integrates object-specific attributes. Specifically, a novel Class-Angle Guidance Module (CAGM) and corresponding loss functions are designed. By incorporating keypoint information, including categories and geometric features, as conditional inputs during the generation process, the proposed framework achieves high-fidelity SAR-to-optical image translation for aircraft targets.

\begin{figure}[tb]
  \centering
\includegraphics[width=\linewidth]{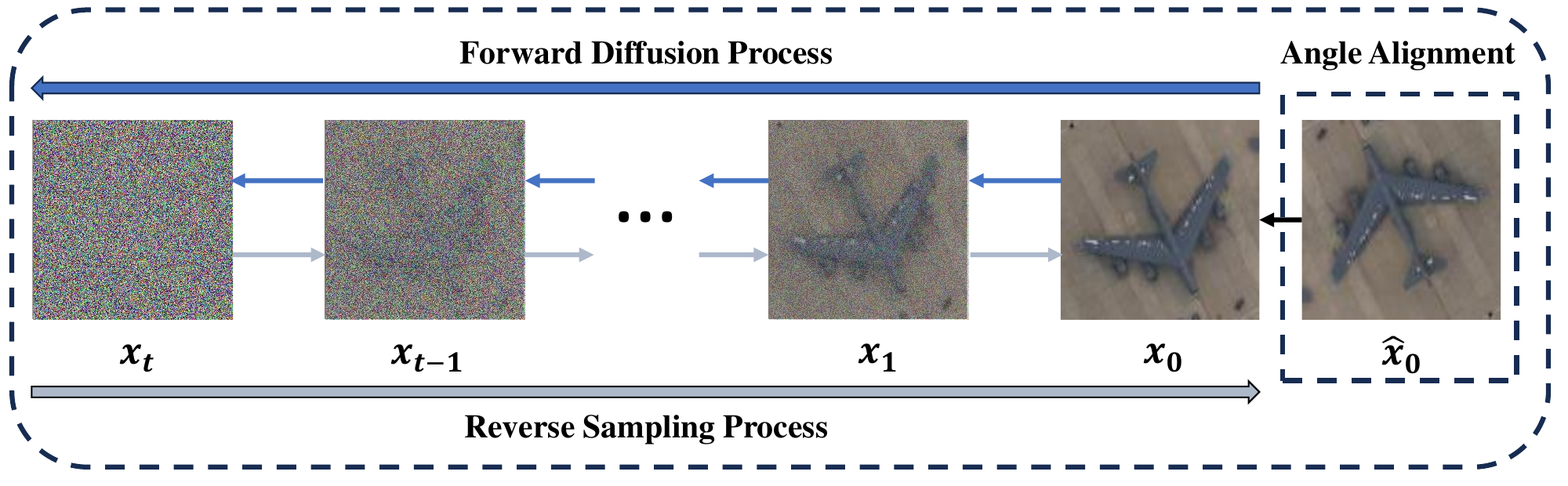}
\caption{Illustrating the forward diffusion process and reverse sampling process. During training, the angle alignment for unpaired training is followed by the forward diffusion process, where the clear image \(x_0\) is gradually diffused by adding the Gaussian noise. In the reverse sampling process, the clear image \(x_0\) is iteratively denoised from an isotropic Gaussian distribution.}
  \label{fig:process}
\end{figure}

\section{Methodology}\label{sec:method}

\subsection{Task Definition}
Given a collection of SAR slice images \( c^{(i,j)} \) and optical slice images \( x^{(i,k)} \) corresponding to aircraft targets, where:\( i \in \{1, \dots, N_{\text{categories}}\} \) denotes the target category,\( j \in \{1, \dots, N_{\text{SAR}}^{(i)}\} \) and \( k \in \{1, \dots, N_{\text{OPT}}^{(i)}\} \) represent the indices of images within the respective modalities for category \( i \),\( N_{\text{SAR}}^{(i)} \) and \( N_{\text{OPT}}^{(i)} \) are the number of SAR and optical images for category \( i \), respectively.

It is important to note that there is no strict spatiotemporal correspondence between the SAR and optical slices. The dataset only provides category labels \( i \) and keypoint annotations \( z \) of each image.

The goal of this study is to develop an automated translation framework capable of converting any given SAR slice image \( c^{(i,j)} \) into its corresponding optical slice image \( x^{(i,k)} \), even in the absence of both category labels and keypoint annotations during inference.

To address this challenging task, this work proposes a novel framework that incorporates domain-invariant keypoint supervision during training, ensuring high-fidelity translation of SAR images to the optical domain while preserving target-specific structural details and semantic consistency.

\subsection{Preliminaries}

\paragraph{Diffusion Model}  
Diffusion models learn a data distribution \( p(\mathbf{x}_0) \) by modeling a forward process that gradually adds noise to data and a reverse process that denoises the noisy data. The forward process is defined as:
\begin{equation}
q(\mathbf{x}_t | \mathbf{x}_{t-1}) = \mathcal{N}(\mathbf{x}_t; \sqrt{1 - \beta_t} \mathbf{x}_{t-1}, \beta_t \mathbf{I}),
\end{equation}
where \( \beta_t \in (0, 1) \) is a variance schedule, and \( \mathbf{x}_t \) represents the noisy version of data \( \mathbf{x}_0 \) at time step \( t \). The forward process allows direct sampling of \( \mathbf{x}_t \) from \( \mathbf{x}_0 \) using:
\begin{equation}
q(\mathbf{x}_t | \mathbf{x}_0) = \mathcal{N}(\mathbf{x}_t; \sqrt{\bar{\alpha}_t} \mathbf{x}_0, (1 - \bar{\alpha}_t) \mathbf{I}),
\end{equation}
where \( \bar{\alpha}_t = \prod_{s=1}^t (1 - \beta_s) \).

The reverse process is defined as:
\begin{equation}
p_\theta(\mathbf{x}_{t-1} | \mathbf{x}_t) = \mathcal{N}(\mathbf{x}_{t-1}; \boldsymbol{\mu}_\theta(\mathbf{x}_t, t), \Sigma_\theta(\mathbf{x}_t, t)).
\end{equation}

In noise prediction parameterization, the model \( \epsilon_\theta(\mathbf{x}_t, t) \) predicts the added noise \( \epsilon \), and the training objective is:
\begin{equation}
\mathcal{L} = \mathbb{E}_{\mathbf{x}_0, \epsilon, t} \left[ \| \epsilon - \epsilon_\theta(\mathbf{x}_t, t) \|^2 \right],
\end{equation}
where \( \mathbf{x}_t = \sqrt{\bar{\alpha}_t} \mathbf{x}_0 + \sqrt{1 - \bar{\alpha}_t} \epsilon \).

\paragraph{Conditional Diffusion Model}  
To incorporate conditions \( \mathbf{c} \) (e.g., images, class labels or keypoint annotations) into the generation process, conditional diffusion models modify the reverse process:
\begin{equation}
p_\theta(\mathbf{x}_{t-1} | \mathbf{x}_t, \mathbf{c}) = \mathcal{N}(\mathbf{x}_{t-1}; \boldsymbol{\mu}_\theta(\mathbf{x}_t, t, \mathbf{c}), \Sigma_\theta(\mathbf{x}_t, t, \mathbf{c})).
\end{equation}

Under noise prediction parameterization, the loss function becomes:
\begin{equation}
\label{equ:noise}
\mathcal{L} = \mathbb{E}_{\mathbf{x}_0, \epsilon, t, \mathbf{c}} \left[ \| \epsilon - \epsilon_\theta(\mathbf{x}_t, t, \mathbf{c}) \|^2 \right].
\end{equation}
This enables the model to leverage semantic or structural information (\( \mathbf{c} \)) during training and generation.

\paragraph{Classifier-Guided Diffusion Model}  
Classifier-guided diffusion improves the conditional diffusion process by using a pretrained classifier \( p_\phi(\mathbf{c} | \mathbf{x}_t) \) to compute the gradient of the log-probability of the condition:
\begin{equation}
\nabla_{\mathbf{x}_t} \log p(\mathbf{x}_t | \mathbf{c}) = \nabla_{\mathbf{x}_t} \log p(\mathbf{c} | \mathbf{x}_t) + \nabla_{\mathbf{x}_t} \log p(\mathbf{x}_t).
\end{equation}
This guidance modifies the reverse process:
\begin{equation}
p_\theta(\mathbf{x}_{t-1} | \mathbf{x}_t, \mathbf{c}) \propto p_\theta(\mathbf{x}_t | \mathbf{c}) \cdot p_\theta(\mathbf{x}_{t-1} | \mathbf{x}_t).
\end{equation}

Classifier guidance enhances fidelity and consistency of generated images but introduces additional computational cost for evaluating the classifier during sampling.

\paragraph{Classifier-Free Guidance Diffusion Model}  
Classifier-free guidance eliminates the need for a separate classifier by training the diffusion model to jointly learn conditional \( \epsilon_\theta(\mathbf{x}_t, t, \mathbf{c}) \) and unconditional \( \epsilon_\theta(\mathbf{x}_t, t) \) denoising objectives. The guided reverse process is then expressed as:
\begin{equation}
\epsilon_\text{guided}(\mathbf{x}_t, t, \mathbf{c}) = (1 + w) \epsilon_\theta(\mathbf{x}_t, t, \mathbf{c}) - w \epsilon_\theta(\mathbf{x}_t, t),
\end{equation}
where \( w \) is the guidance scale that controls the trade-off between fidelity (conditional consistency) and diversity (unconditional generative quality).

The corresponding training objective remains:
\begin{equation}
\mathcal{L} = \mathbb{E}_{\mathbf{x}_0, \epsilon, t, \mathbf{c}} \left[ \| \epsilon - \epsilon_\text{guided}(\mathbf{x}_t, t, \mathbf{c}) \|^2 \right].
\end{equation}

Classifier-free guidance provides an efficient alternative while maintaining high-fidelity conditional generation, making it suitable for object-level unpaired translation tasks with high-generation possibility.

\subsection{Framework Overview}

Fig.~\ref{fig:overview} illustrates the overview of KeypointDiff. The proposed framework is built upon the Classifier-Free Guidance (CFG) diffusion model, tailored for high-fidelity SAR-to-OPT image translation for aircraft with precise structural alignment. It integrates a keypoint-driven training strategy that utilizes annotated keypoints to supervise the preservation of complex structural details during translation, an automated testing workflow to eliminate dependency on annotations during inference, and the CAGM (Class-Angle Guidance Module), which encodes aircraft-specific attributes such as class and orientation into the diffusion process. Additionally, customized loss functions ensure image fidelity and structural consistency, addressing the challenges of unpaired image translation.

By combining these components, the framework effectively bridges the domain gap between SAR and OPT images, generating high-quality OPT images with accurate structural and semantic alignment. This holistic design offers a robust and scalable solution for the challenging task of unpaired SAR-to-OPT image translation.

\subsection{Keypoints-based Training and Testing Strategies}

As shown in Fig.~\ref{fig:process}, the proposed framework incorporates both the diffusion process during training and the sampling process during testing to effectively translate SAR images to optical images. The diffusion process involves gradually adding noise to the image during training, followed by denoising to produce high-fidelity outputs. In contrast, the sampling process during testing recovers the clean optical image from the noisy input, guiding the model to perform image translation. Throughout both training and testing, a keypoints-based supervision strategy is employed to ensure the alignment of target features, enabling more precise generation.

\begin{figure}[tb]
  \centering
\includegraphics[width=\linewidth]{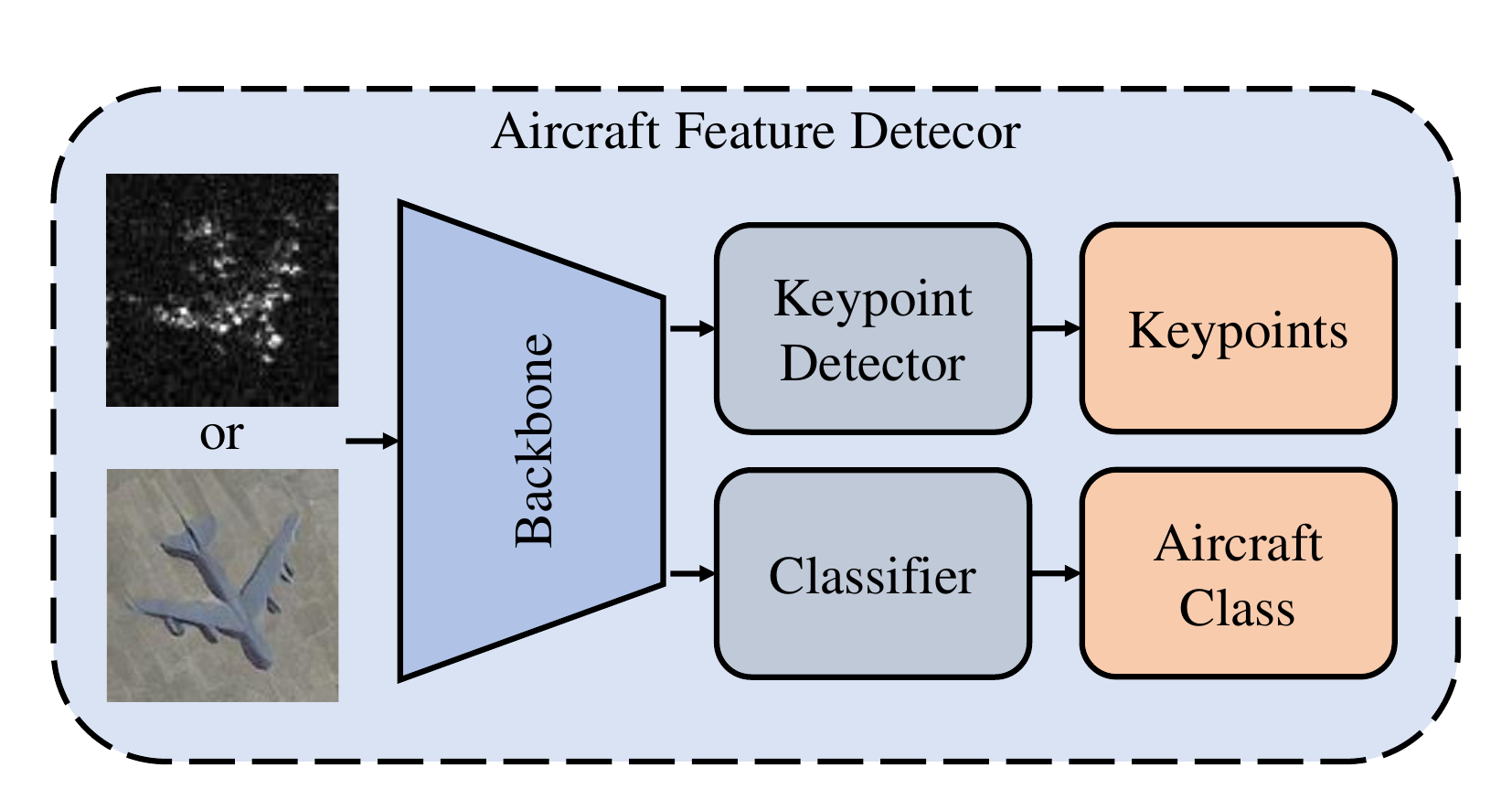}
\caption{Structure of the aircraft feature detector predicting aircraft keypoints and the category.}
  \label{fig:detector}
\end{figure}

Keypoints-based supervision for aircraft is implemented through the training of two aircraft feature detectors, one for optical images and another for SAR images, as shown in Fig~\ref{fig:detector}. The detector backbone consists of a ResNet50 and a 3-layer Feature Pyramid Networks (FPN), followed by a heatmap-based keypoint detector inspired by \cite{heatmap}, and a classifier. The keypoint detection module identifies two critical points of the aircraft nose and tail, while the classifier predicts the category of the image. Both the training and testing datasets for the detectors are consistent with those used in the subsequent SAR-to-OPT translation task, and the pre-trained detector results are summarized in Table~\ref{tab:detector}.

\paragraph{Training}  
In the absence of paired data, a pseudo-pairing strategy is employed, driven by keypoint annotations and dynamic optimization. A dynamic pairing approach selects structurally similar optical slices for each SAR slice. To address angular discrepancies, a rotation function preprocesses the optical images, aligning their orientation with the SAR images based on their respective keypoints as shown in Fig~\ref{fig:process}. During this rotation, padding is applied to the optical images' background to fill in the areas that are not aligned, considering that the aircraft slices typically have a uniform background such as a runway, making such padding acceptable. This angle alignment improves geometric consistency and ensures better initialization of pseudo-pairs. These pairs are iteratively optimized during training, refining the distribution alignment between SAR and optical domains and progressively improving the model's translation performance.

\paragraph{Testing}  
During inference, the framework enables fully automated translation of SAR images without relying on explicit category or angle annotations. A pre-trained detector for SAR aircraft predicts approximate keypoints and target categories, which are embedded as conditional vectors into the model. These embeddings serve as implicit supervision, allowing the model to generate high-fidelity optical images from SAR inputs without requiring additional metadata during testing.

\begin{table}
\begin{center}
\caption{Performance of Pre-trained Detectors}
\label{tab:detector}
\begin{tabular}{ccc}
\toprule
Domain & OA(↑) & Angle Error(↓) \\
\midrule
 SAR & 0.8283 & 20.7747\\
 OPT & 0.9774 & 4.7324\\
\bottomrule
\end{tabular}
\end{center}
\end{table}

This dual strategy, combining keypoint-guided pseudo-pairing during training with automated keypoint embedding during testing, forms a robust foundation for the SAR-to-OPT image translation pipeline. By effectively addressing the challenges posed by unpaired datasets and intricate target structures, this framework ensures accurate and reliable translations even in the presence of complex data.

\subsection{Class-Angle Guidance Module}

\begin{figure}[tb]
  \centering
\includegraphics[width=\linewidth]{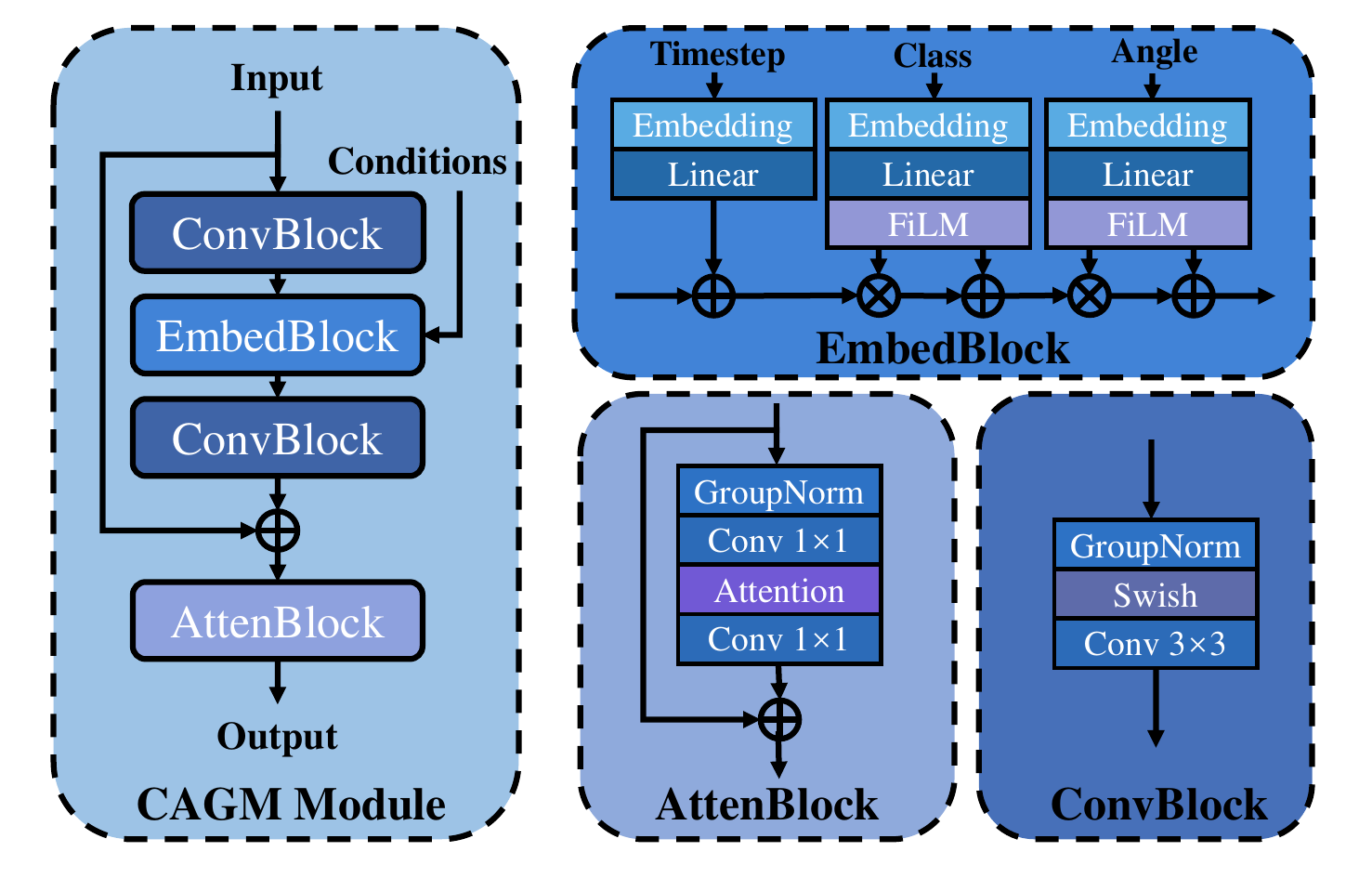}
\caption{Structure of the CAGM module and its sub-blocks, forming the core component of the denoising U-Net in the KeypointDiff framework.}
  \label{fig:module}
\end{figure}

The Class-Angle Guidance Module (CAGM) serves as a critical building block within the denoising U-Net of KeypointDiff, effectively integrating conditional information including aircraft class, and angle into the denoising process. The module is composed of three key components as illustrated in Fig.~\ref{fig:module}: ConvBlock, AttentionBlock, and EmbedBlock, which work together to facilitate local feature extraction, global dependency modeling, and conditional embedding.

The ConvBlock is primarily responsible for feature extraction and refinement. It standardizes the input using group normalization, introduces non-linearity through a Swish activation function, and applies a \(3 \times 3\) convolution to extract spatial features. The computation can be represented as:
\begin{equation}
\mathbf{f}_{\text{Conv}}(\mathbf{x}) = \text{Conv}\left( \text{Swish} \left( \text{GroupNorm}(\mathbf{x}) \right) \right),
\end{equation}
where \(\mathbf{x}\) denotes the input features, and \(\mathbf{f}_{\text{Conv}}(\mathbf{x})\) represents the refined output.

The AttentionBlock is designed to capture long-range dependencies via a self-attention mechanism~\cite{attention}. First, the input undergoes group normalization, followed by a \(1 \times 1\) convolution to project the features into query (\(\mathbf{q}\)), key (\(\mathbf{k}\)), and value (\(\mathbf{v}\)) matrices. The self-attention mechanism computes:
\begin{equation}
\text{Attention}(\mathbf{q}, \mathbf{k}, \mathbf{v}) = \text{softmax} \left( \frac{\mathbf{q}\mathbf{k}^\top}{\sqrt{d}} \right) \mathbf{v},
\end{equation}
where \(d\) is the dimensionality of the feature space. The output is then projected back into the original feature space using another \(1 \times 1\) convolution. A residual connection is applied to add the processed attention output to the original input:
\begin{equation}
\mathbf{f}_{\text{Attn}}(\mathbf{x}) = \mathbf{x} + \text{Conv} \left( \text{Attention}(\mathbf{q}, \mathbf{k}, \mathbf{v}) \right).
\end{equation}
This process allows the model to encode both local and global information effectively.

% The EmbedBlock is responsible for embedding conditional information such as timestep, class, and angle into the feature space. The timestep information is first embedded through a learned positional embedding and transformed via a linear layer, then directly added to the input feature. For class and angle information, embeddings are generated through learned embedding layers, followed by linear transformations. These embeddings are then used in a Feature-wise Linear Modulation (FiLM) operation\cite{FiLM} to compute shift (\(\boldsymbol{\gamma}\)) and scale (\(\boldsymbol{\beta}\)) parameters, which modulate the input features as:
% \begin{equation}
% \text{FiLM}(\mathbf{x}) = \boldsymbol{\gamma} \cdot \mathbf{x} + \boldsymbol{\beta},
% \end{equation}
% where \(\boldsymbol{\gamma}\) and \(\boldsymbol{\beta}\) are functions of the conditional embeddings, and \(\mathbf{x}\) represents the input features.This design allows the model to seamlessly incorporate semantic and structural information into the generation process.

The EmbedBlock is the core component for encoding aircraft features. It sequentially integrates timestep and aircraft feature information including class and angle embeddings into the feature space through a Feature-wise Linear Modulation (FiLM) operation~\cite{FiLM}, which modulates features using condition-dependent shift (\(\boldsymbol{\beta}\)) and scale (\(\boldsymbol{\gamma}\)) parameters:
\begin{equation}
\text{FiLM}(\mathbf{x}) = \boldsymbol{\gamma} \cdot \mathbf{x} + \boldsymbol{\beta},
\end{equation}
where \(\boldsymbol{\gamma}\) and \(\boldsymbol{\beta}\) are learned from the conditional embeddings through a linear layer. FiLM effectively allows the model to encode semantic and structural information by selectively amplifying or attenuating input features.

The EmbedBlock sequentially encodes timestep, class, and angle information into the feature space:
First, the input timestep is passed through a learned embedding and transformed via a linear layer. This embedding is then added directly to the input feature:
\begin{equation}
\mathbf{f}_{\text{Timestep}}(\mathbf{x}) = \mathbf{x} + \text{Linear}(\text{Embed}(\text{Timestep})).
\end{equation}

Then, the class information is embedded and transformed linearly. The resulting parameters (\(\boldsymbol{\gamma}_\text{class}\) and \(\boldsymbol{\beta}_\text{class}\)) are computed for FiLM modulation:
\begin{equation}
\mathbf{f}_{\text{Class}}(\mathbf{x}) = \boldsymbol{\gamma}_\text{class} \cdot \mathbf{f}_{\text{Timestep}}(\mathbf{x}) + \boldsymbol{\beta}_\text{class}.
\end{equation}

Similarly, angle information is embedded to compute the parameters (\(\boldsymbol{\gamma}_\text{angle}\) and \(\boldsymbol{\beta}_\text{angle}\)) for FiLM modulation:
\begin{equation}
\mathbf{f}_{\text{Angle}}(\mathbf{x}) = \boldsymbol{\gamma}_\text{angle} \cdot \mathbf{f}_{\text{Class}}(\mathbf{x}) + \boldsymbol{\beta}_\text{angle}.
\end{equation}

The final output of the EmbedBlock is the encoded feature \(\mathbf{f}_{\text{Embed}}(\mathbf{x})\), which integrates timestep, class, and angle information:
\begin{equation}
\mathbf{f}_{\text{Embed}}(\mathbf{x}) = \mathbf{f}_{\text{Class}}(\mathbf{f}_{\text{Class}}(\mathbf{f}_{\text{Timestep}}(\mathbf{x}))).
\end{equation}

The overall structure of the CAGM combines these three blocks in a coherent manner. The module begins with a ConvBlock for initial feature extraction, followed by an EmbedBlock to encode conditional information. Another ConvBlock is then applied to refine the features, and a residual connection is maintained throughout the module. Finally, an AttentionBlock is used to capture global dependencies. The entire operation of the CAGM can be summarized as:
\begin{equation}
\mathbf{y} = \mathbf{f}_{\text{Attn}} \left( \mathbf{x} + \mathbf{f}_{\text{Conv}} \left( \mathbf{f}_{\text{Embed}} \left( \mathbf{f}_{\text{Conv}}(\mathbf{x}) \right) \right) \right),
\end{equation}
where \(\mathbf{x}\) is the input, and \(\mathbf{y}\) is the output. By combining local feature extraction, global dependency modeling, and conditional embedding, the CAGM effectively guides the denoising U-Net to produce high-fidelity optical images from SAR inputs, preserving both structural details and semantic consistency.

\subsection{Joint Loss Functions}

The proposed framework optimizes a joint loss function to ensure high-fidelity translation and accurate target-specific feature preservation. The joint loss function consists of three components, each targeting a specific aspect of the SAR-to-OPT translation task.

\paragraph{Base Noise Prediction Loss (\(\mathcal{L}_\text{simple}\))}  
The foundational loss of the diffusion model is the base noise prediction loss, \(\mathcal{L}_\text{simple}\). This loss minimizes the discrepancy between the predicted noise \(\epsilon_\theta(\mathbf{x}_t,c,t,i,z)\) of SAR image condition \(c \) and feature conditions of class \(i\) and keypoints-based angle \( z \) and the true noise \(\epsilon\) added during the forward diffusion process, formulated as:
\begin{equation}
\mathcal{L}_\text{simple} = \mathbb{E}_{\mathbf{x}_t, \epsilon, t} \left[ \| \epsilon - \epsilon_\theta(\mathbf{x}_t,c,t,i,z) \|^2 \right].
\end{equation}
This ensures that the model effectively denoises and reconstructs high-quality images during the reverse process.

\paragraph{Image Consistency Loss (\(\mathcal{L}_\text{consistency}\))}  
To maintain the consistency and realism of the generated images, an image consistency loss \(\mathcal{L}_\text{consistency}\) is employed. The consistency loss is given as:
\begin{equation}
\mathcal{L}_\text{consistency} = \mathcal{L}_\text{color} + \mathcal{L}_\text{percep}.
\end{equation}  

\textit{Color Loss (\(\mathcal{L}_\text{color}\)):} This computes the similarity between the RGB vectors of the generated image \(\mathbf{x}_\text{pred}\) through one-step prediction and the target image \(\mathbf{x}\). It encourages alignment of the color distribution, defined as:
\begin{equation}
\mathcal{L}_\text{color} = 1 - \frac{1}{N} \sum_{i=1}^N \left( \frac{\mathbf{x}^\text{pred}_\text{} \cdot \mathbf{x}^\text{opt}}{\|\mathbf{x}^\text{pred}_\text{}\| \|\mathbf{x}^\text{opt}\|} \right),
\end{equation}
where \(N\) is the number of color channels (3 for RGB here).

\textit{Perceptual Loss (\(\mathcal{L}_\text{percep}\)):} This utilizes a pre-trained VGG network to extract deep feature representations of both the generated and target images\cite{perceptual_loss}. The MSE is then calculated between the feature vectors:
\begin{equation}
\mathcal{L}_\text{percep} = \mathbb{E} \left[ \| \phi_\text{VGG}(\mathbf{x}^\text{pred}) - \phi_\text{VGG}(\mathbf{x}^\text{opt}) \|^2 \right].
\end{equation}

These two terms jointly ensure both low-level and high-level consistency between the generated and target images.

\paragraph{Adversarial Loss (\(\mathcal{L}_\text{adv}\))}  
To accurately preserve the target's category-specific features, an adversarial loss \(\mathcal{L}_\text{adversarial}\) is introduced. A pre-trained optical image detector is employed to predict the category vector and keypoint heatmaps of the generated image. The loss consists of two components:

\textit{Category Loss:} Cross-entropy loss is applied to compare the predicted and ground-truth category vectors.
\begin{equation}
\mathcal{L}_\text{category} = - \sum_{i=1}^{N} \mathbf{y}_\text{i}^{SAR} \log(\mathbf{y}_\text{i}^{pred}),
\end{equation}
where \(N\) is the number of categories, \(\mathbf{y}_\text{i}^{SAR}\) is the true category label in one-hot encoding, and \(\mathbf{y}_\text{i}^{pred}\) is the predicted category probability.

\textit{Keypoint Heatmap Loss:} MSE is used to compare the predicted and ground-truth keypoint heatmaps.
\begin{equation}
\mathcal{L}_\text{keypoints} = \frac{1}{N} \sum_{i=1}^{N} \|\mathbf{H}_\text{i}^{pred} - \mathbf{H}_\text{i}^{SAR}\|_2^2,
\end{equation}
where \(N\) is the number of pixels in the heatmap, and \(\mathbf{H}_\text{i}^{pred}\) and \(\mathbf{H}_\text{i}^{SAR}\) represent the predicted and ground-truth keypoint heatmaps, respectively.

The adversarial loss is given as:
\begin{equation}
\mathcal{L}_\text{adv} = \mathcal{L}_\text{category} + \mathcal{L}_\text{keypoints}.
\end{equation}

\paragraph{Combined Loss}  
The total loss combines all three components as follows:
\begin{equation}
\mathcal{L}_\text{total} = \mathcal{L}_\text{simple} + \lambda_\text{consistency} \mathcal{L}_\text{consistency} + \lambda_\text{adv} \mathcal{L}_\text{adv},
\end{equation}
where \(\lambda_\text{consistency}\) and \(\lambda_\text{adv}\) are hyperparameters to balance the contributions of each term. This joint loss ensures high-fidelity image generation, structural consistency, and accurate preservation of target-specific details in the translation task.

\section{Experiments}
\label{sec:experiment}

\subsection{Experiment Setup}

\begin{table}[tb]
\begin{center}
\caption{Summary of Dataset Statistics}
\label{tab:dataset}
\resizebox{\linewidth}{!}{
\begin{tabular}{ccccc}
\toprule
Categories & SAR Train & SAR Test & OPT Train & OPT Test \\
\midrule
15 & 2,172 & 99 & 2,583 & 133 \\
\bottomrule
\end{tabular}
}
\end{center}
\end{table}

\begin{figure}[tb]
  \centering
\includegraphics[width=\linewidth]{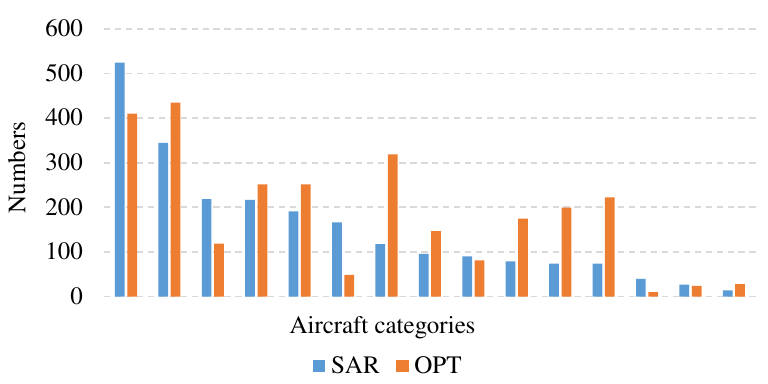}
\caption{Overview of the dataset used in the experiments, showing the number of SAR and optical images for each aircraft category.}
  \label{fig:dataset}
\end{figure}

\paragraph{Dataset Description}  

The SAR images utilized in the experiments are sourced from the Gaofen-3 satellite, focusing on aircraft target slices with a spatial resolution of 1 meter. Optical images are derived from the dataset presented in \cite{yutong_qian}, as well as additional data obtained from the \textit{Bigemap} platform, featuring resolutions ranging from 0.5 to 1 meter. Both SAR and optical images are preprocessed into aircraft target slices to ensure spatial consistency and comparable regions for translation tasks. This alignment facilitates robust evaluation of the SAR-to-optical image translation methods presented in this study. Table~\ref{tab:dataset} summarizes the number of training and testing samples for both the SAR and optical images across 15 categories. Fig.~\ref{fig:dataset} provides a visual overview of the dataset composition, displaying the number of SAR and optical images for each category.

\paragraph{Metrics}  
The evaluation of the proposed method is conducted from two perspectives:  
1. \textit{Image Generation Quality}: The Fréchet Inception Distance (FID)\cite{FID} is used to assess the visual fidelity and diversity of the generated optical images.  
2. \textit{Aircraft-Specific Features}: A pre-trained aircraft detector is employed to evaluate the generated optical images based on two metrics:  
Overall classification accuracy (OA), which measures the correctness of aircraft type classification.  
Angle Detection Error (Angle Error), which quantifies the discrepancy between the detected and ground-truth orientation angles of the aircraft.  

\paragraph{Implementation Details}  
For all experiments, the input images are normalized and resized to a uniform resolution of $128 \times 128$. Data augmentation is applied during training, including random horizontal flips (50\%), random vertical flips (50\%), and random rotations with a probability of 50\% (rotation angle within $\pm 30$ degrees).

The model is trained on four NVIDIA RTX 4090 GPUs using the AdamW optimizer with a learning rate of $1 \times 10^{-4}$ and a weight decay of $1 \times 10^{-4}$ for 400,000 iterations. The total time-step of the diffusion process $T$ is set to 1,000. The batch size is set to 16. To adjust the learning rate, a cosine annealing scheduler~\cite{cosine_lr} is employed. During the initial 4,000 iterations, a warm-up scheduler is applied to gradually increase the learning rate. Subsequently, $\mathcal{L}_\text{consistency}$ and $\mathcal{L}_\text{adv}$ begin to contribute, with $\lambda_\text{consistency}$ and $\lambda_\text{adv}$ set to 0.01 and 0.001, respectively.

During testing, the SAR input images are first processed using a pre-trained SAR aircraft feature detector to extract approximate aircraft attributes, including category and orientation. These limited attributes are then utilized to guide the translation process. The guidance scale $w$ is set to 1.

\paragraph{Comparison with Other Methods}  
The proposed approach is evaluated against the following baseline methods:  
\textit{Simply Rotation}: For each SAR input image, a randomly selected optical image from the training set belonging to the detected aircraft category is chosen. The optical image is rotated to align with the input SAR image based on their angle differences.  
\textit{CycleGAN} \cite{unpaired_cyclegan}: A classic GAN-based unpaired image-to-image translation framework.  
\textit{Contrastive Unpaired Translation (CUT)} \cite{unpaired_CUT}: A classic contrastive learning-based unpaired image-to-image translation method.  
\textit{UNSB} \cite{unpaired_UNSB}: A Schrödinger Bridge-based diffusion model for unpaired image translation.  
\textit{Guided Diffusion} \cite{diffusion_guided-diffusion}: A diffusion model guided by a pre-trained classifier. The detected category serves as the initial condition to guide the diffusion process.  

\begin{figure*}[tb]
  \centering
\includegraphics[width=\linewidth]{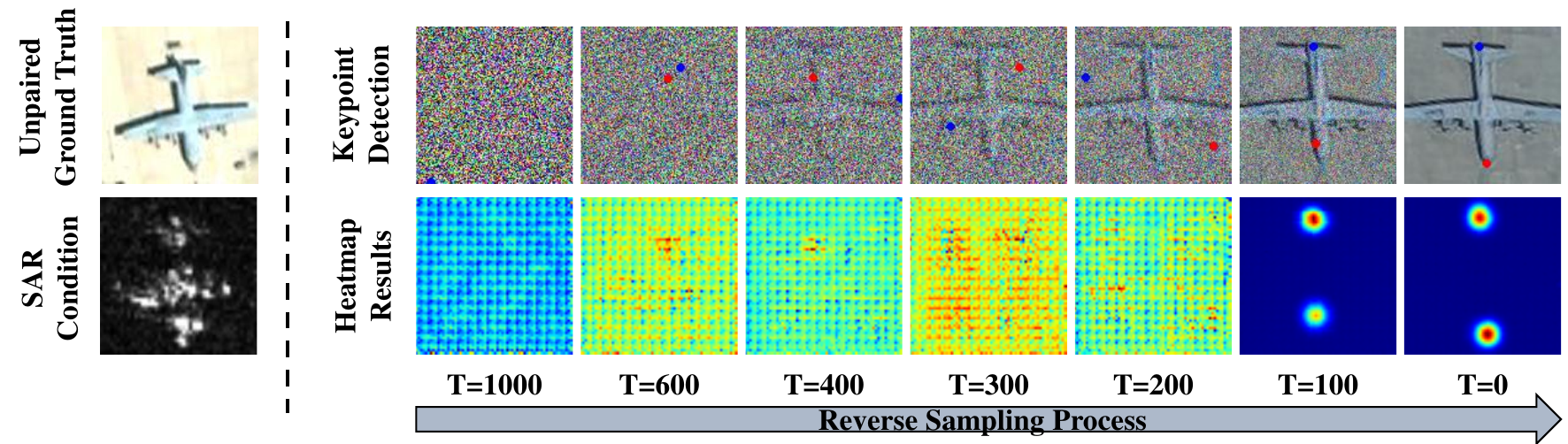}
\caption{Denoising process and heatmap-based keypoint detection results.}
  \label{fig:heatmap}
\end{figure*}

Fig~\ref{fig:heatmap} illustrates the denoising process, demonstrating how the generated image evolves from random noise to the corresponding optical image. The progression reveals increasingly clear and accurate representations of the target. Additionally, the heatmap-based keypoint detection results indicate that the structural clarity of keypoints, such as nose and tail, improves significantly during the denoising process. It is important to note that the unpaired ground truth only represents manually selected optical aircraft images of the same category as the SAR images, with appropriate rotation applied for comparison, and does not represent spatiotemporal alignment.

\subsection{Experimental Results}
\paragraph{Quantitative Results} 

% \begin{table}[tb]
% \begin{center}
% \caption{Ablation studies on components of KeypointDiff}
% \label{tab:ablation}
% \resizebox{\columnwidth}{!}{
% \begin{tabular}{cccc}
% \toprule
% Methods & FID↓ & Ang. Error↓ & OA↑ \\
% \midrule
% Baseline & 223.5673 & 51.3790 & 0.1314\\
% +Keypoints-based Alignment & 223.5673 & 51.3790 & 0.1314\\
% +Keypoints-based Alignment +CAGM & 165.5375  & 22.7159 & 0.7071\\
% +Keypoints-based Alignment +CAGM +\(\mathcal{L}_\text{color}\) & 159.2050 & 23.5476 & 0.6869\\
% +Keypoints-based Alignment +CAGM +\(\mathcal{L}_\text{color}\) +\(\mathcal{L}_\text{percep}\) & 148.1736 & 22.1080 & 0.6970\\
% +Keypoints-based Alignment +CAGM +\(\mathcal{L}_\text{color}\) +\(\mathcal{L}_\text{percep}\) + \(\mathcal{L}_\text{adv}\) & \textbf{140.7208} & \textbf{21.5970} & \textbf{0.7374}\\
% \bottomrule
% \end{tabular}
% }
% \end{center}
% \end{table}

As shown in Table~\ref{tab:comparison}, the proposed method outperforms domain-shift-based methods and classifier-guided diffusion using the proposed keypoints-based training strategy. Specifically, compared to domain-shift-based techniques, which focus on aligning the entire image distribution between SAR and optical domains, the proposed method enhances translation performance via the proposed keypoint-based guidance. This approach allows the model to align the structural details of the targets more precisely, as it incorporates target-specific spatial information. 
The tailored network design and loss functions further improve performance by preserving category-specific details and learning domain-invariant features, ensuring visually accurate and semantically coherent translations. This enables finer structural preservation in SAR-to-optical translation, overcoming the limitations of domain-shift methods in recovering intricate details.

While the simply rotation method, which directly rotates the optical image based on the detected class and orientation, guarantees high fidelity in real optical images from the view of FID scores, it heavily relies on accurate predictions of the category and orientation, which limits its generalization ability. Despite this, the overall accuracy (OA) of the proposed method strikes a balance between precision and generalization, addressing the need to align different categories while improving downstream task performance, which will be discussed further in \ref{sec:disscusion}. 

In terms of angle accuracy (Angle Error), the proposed method outperforms others because it integrates the detected orientation information into the translation process. By combining the SAR image with the predicted orientation, the model can accurately generate the corresponding optical image with well-aligned target orientations, ensuring better angle accuracy than the simpler rotation or domain-shift methods.

\begin{figure*}[htb]
  \centering
\includegraphics[width=\linewidth]{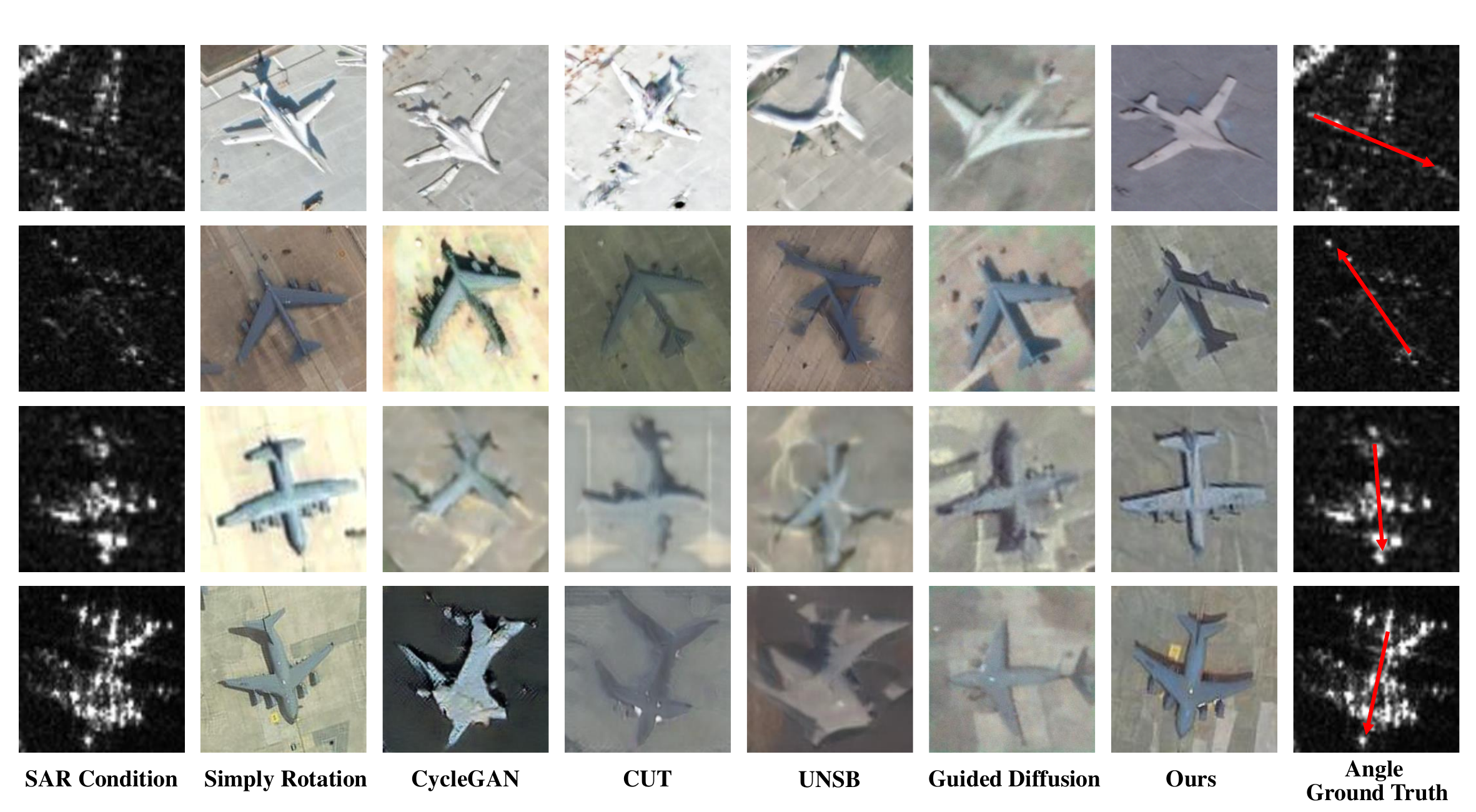}
\caption{Qualitative comparison of the proposed method with other approaches, highlighting differences in image quality and target details.}
  \label{fig:comparison}
\end{figure*}

\begin{table}[tb]
\begin{center}
\caption{Performance comparisons with other methods}
\label{tab:comparison}
\resizebox{\columnwidth}{!}{
\begin{tabular}{cccc}
\toprule
Methods & FID↓ & Angle Error↓ & OA↑ \\
\midrule
Simply Rotation & 117.7661 & 23.0046 & 0.7980\\
\midrule
CycleGAN & 246.5647 & 72.2578 & 0.5556\\
CUT & 234.5614 & 71.5156 & 0.6667\\
UNSB & 228.3685 & 75.6435 & 0.6768\\
Guided Diffusion& 206.9595 & 61.7034 & 0.7071\\
Ours & \textbf{140.7208} & \textbf{21.5970} & \textbf{0.7374}\\
\bottomrule
\end{tabular}
}
\end{center}
\end{table}

\paragraph{Qualitative Results} 

Fig~\ref{fig:comparison} compares the visual results of the proposed method with other approaches. 
The selected results of the simply rotation method exhibits real optical images with correct categories; however, the angles are misaligned due to the lack of fine-grained spatial alignment between the SAR and optical domains. 

On the other hand, domain-shift-based methods, such as CycleGAN and CUT, fail to restore category-specific contour and structural information accurately. These methods rely on aligning the general distribution between domains, which can lead to significant blurring and loss of structural details, especially for target-specific features like contours, wings, or other aircraft components. 

Classifier-guided diffusion performs relatively well by incorporating pre-trained classifiers to guide the generation process. While this improves the structural mapping of targets, the lack of task-specific design means that the alignment between the generated category and angle remains suboptimal. This results in a less accurate representation of target features, such as shape and texture. 

In contrast, the proposed method effectively integrates keypoint guidance and category-angle embedding, leveraging both the SAR image and predicted category and angle information for enhanced translation. By doing so, the method achieves superior results in both fidelity and structural consistency . These results demonstrate that the proposed approach, which uses keypoints for alignment and embedding for accurate category-angle prediction, provides a significant advantage over other methods in preserving structural integrity and generating high-quality optical images.

\subsection{Ablation Study}

\begin{figure*}[htb]
  \centering
\includegraphics[width=\linewidth]{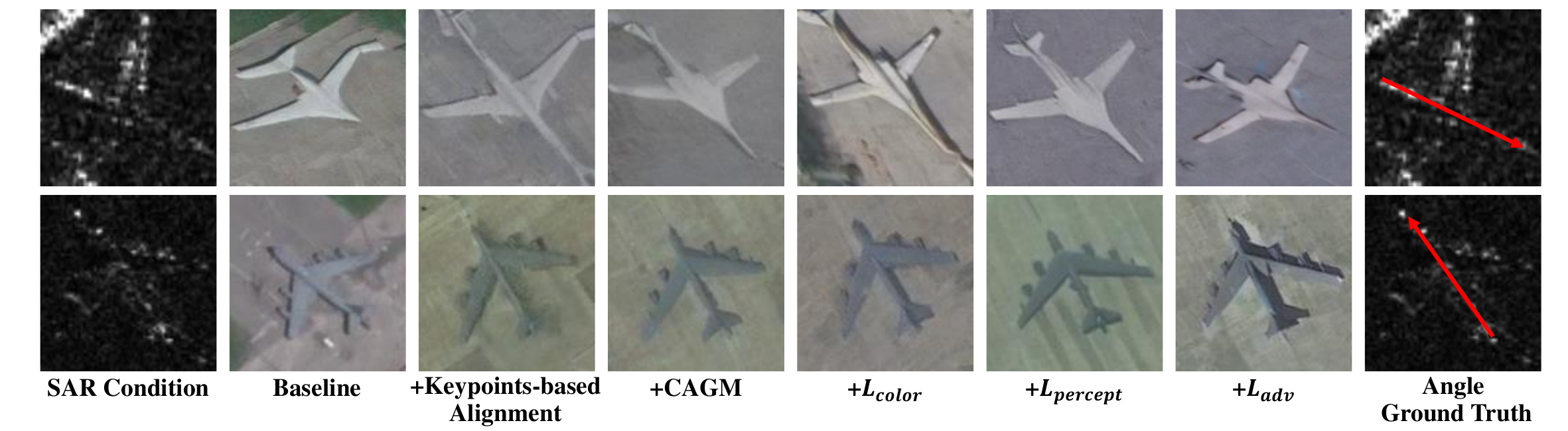}
\caption{Ablation study results, demonstrating the impact of different components on the performance of the proposed method.}

  \label{fig:ablation}
\end{figure*}

Ablation experiments are conducted to evaluate all the individual components in the proposed method. Table~\ref{tab:ablation} quantitatively illustrates the impact of each component, while Figure~\ref{fig:ablation} provides a qualitative comparison of the results.

\begin{table}[tb]
\begin{center}
\caption{Ablation Studies on Components of KeypointDiff}
\label{tab:ablation}
\resizebox{\columnwidth}{!}{
\begin{tabular}{ccccc|ccc}
\toprule
KP-A & CAGM & \(\mathcal{L}_\text{color}\) & \(\mathcal{L}_\text{percep}\) & \(\mathcal{L}_\text{adv}\) & FID↓ & Ang. Error↓ & OA↑ \\
\midrule
  &  &  &  &  & 223.5673 & 51.3790 & 0.3131\\
 \checkmark &  &  &  &  & 181.2403 & 38.1331 & 0.6768\\
 \checkmark & \checkmark &  &  &  & 165.5375 & 22.7159 & 0.7071\\
 \checkmark & \checkmark & \checkmark &  &  & 159.2050 & 23.5476 & 0.6869\\
 \checkmark & \checkmark & \checkmark & \checkmark &  & 148.1736 & 22.1080 & 0.6970\\
 \checkmark & \checkmark & \checkmark & \checkmark & \checkmark & \textbf{140.7208} & \textbf{21.5970} & \textbf{0.7374} \\
\bottomrule
\end{tabular}
}
\end{center}
\end{table}

The baseline model, which implements a standard classifier-free guidance (CFG) approach without the keypoint-based training strategy, represents the most basic version of the proposed framework.  However, without keypoint-based pairing during training, this baseline model struggles to capture important target-specific details, leading to blurry images with poor texture fidelity. The generated optical images often fail to preserve accurate structural details,including the aircraft’s shape and key angles, and they lack proper category alignment. This results in high FID scores and poor visual quality compared to real optical images. 

Introducing the keypoint-based training strategy for alignment, which is marked as KP-A in Table~\ref{tab:ablation}, addresses this limitation by providing spatially aligned training data. By using keypoint annotations, the diffusion model can be guided in capturing the target geometry more accurately. This allows the model to align features in both SAR and OPT domains based on corresponding keypoints, thus improving the generation of aircraft targets to reconstruct structural details. However, the overall image quality is still not optimal, with some blending between the target and background.

Adding the Class-Angle Guidance Module (CAGM) enhances the generation process further by embedding both the aircraft class and angle information into the model. CAGM introduces critical guidance, ensuring that the model accurately generates targets in the correct category and orientation. With the integration of CAGM, the model can produce more accurate representations of target structures. However, despite the improved structural accuracy, the generated images still exhibit some blurry background regions, and there is a slight color shift, which affects overall image fidelity.

The addition of \(\mathcal{L}_\text{color}\) and \(\mathcal{L}_\text{percep}\)further improves the image quality. The color loss ensures that the generated images' color distribution matches that of the real optical images, correcting issues of unnatural color shifts often seen in generated images, while the perceptual loss is designed to enhance the high-level features in the generated image, particularly texture details, contours, and surface structures. The final addition of \(\mathcal{L}_\text{adv}\) plays a crucial role in refining the final output, encouraging the model in a more adversarial manner to generate images that are not only perceptually realistic but also structurally consistent with the target categories and angles.

The visual results in Figure~\ref{fig:ablation} show that the final proposed model produces the most accurate, high-fidelity optical images, with improved texture details, better category alignment, and precise angles. This ablation study demonstrates that the combination of keypoint-based training strategy, CAGM, and customized loss function design is essential for achieving superior performance of the proposed KeypointDiff.

\subsection{Parameter Sensitivity and Generalization Analysis}
\label{sec:disscusion}
\paragraph{Influence of $w$ in Classifier-Free Guidance Sampling} 
% The guidance scale \(w\) significantly affects both the quality and diversity of the generated images. A higher \(w\) increases the influence of conditional control, which can enhance fidelity but may result in unrealistic outputs due to overfitting to the conditions. Table~\ref{tab:w_influence} shows the impact of different \(w\) values on FID, angle error, and OA. Based on these results,  \(w=1\) was selected as the optimal configuration for the experiments, as it achieves the best balance across all evaluation metrics.

\begin{table}[tb]
\begin{center}
\caption{Impact of \(w\) in Classifier-Free Guidance sampling}
\label{tab:w_influence}
\resizebox{\columnwidth}{!}{
\begin{tabular}{lccc}
\toprule
Methods & FID↓ & Angle Error↓ & OA↑ \\
\midrule
$w$=0 & 160.9024 & 23.4834 & 0.5455\\
$w$=0.5 & 146.2224 & \textbf{19.7975} & 0.6465\\
$w$=1 & \textbf{140.7208} & 21.5970 & \textbf{0.7374}\\
$w$=3 & 161.0174 & 22.4427 & 0.5859\\
\bottomrule
\end{tabular}
}
\end{center}
\end{table}

The guidance scale \(w\) plays a crucial role in controlling the influence of the conditional inputs on the generation process. A higher value of \(w\) amplifies the impact of the input conditions, such as category and angle, improving the alignment between the generated image and the desired target attributes. However, this increased control can lead to overfitting, where the generated images become too rigidly tied to the input conditions, potentially sacrificing diversity or introducing unrealistic artifacts. On the other hand, a lower value of \(w\) may reduce overfitting but could result in less accurate image generation, as the conditional inputs play a smaller role in guiding the model.

As shown in Table~\ref{tab:w_influence}, the variation of \(w\) has a notable effect on key evaluation metrics. Specifically, it can be observed that \(w=1\) achieves the comparative balance across all metrics, offering an optimal trade-off between fidelity and diversity. This value of \(w\) ensures that the model generates high-quality images while preserving variation in the outputs, leading to realistic results without overfitting to specific conditions.

\paragraph{Influence of Conditions}  
% The input to the denoising model consists of two key components: the SAR image condition and the feature condition provided by the detector. Figure~\ref{fig:no_sar} illustrates the results of ablation experiments analyzing their individual contributions.  
% When only the feature condition is used, the generated images are heavily influenced by the detector. In cases of low confidence or incorrect angle predictions, this results in blurry and inaccurate target outlines, as well as errors in the target's orientation.  
% Conversely, using only the SAR image without the feature condition leads to inaccuracies in target attributes, such as incorrect engine counts or misidentification of aircraft types.  
% These findings highlight the importance of both conditions in ensuring accurate and realistic image generation.

\begin{figure}[tb]
  \centering
\includegraphics[width=\linewidth]{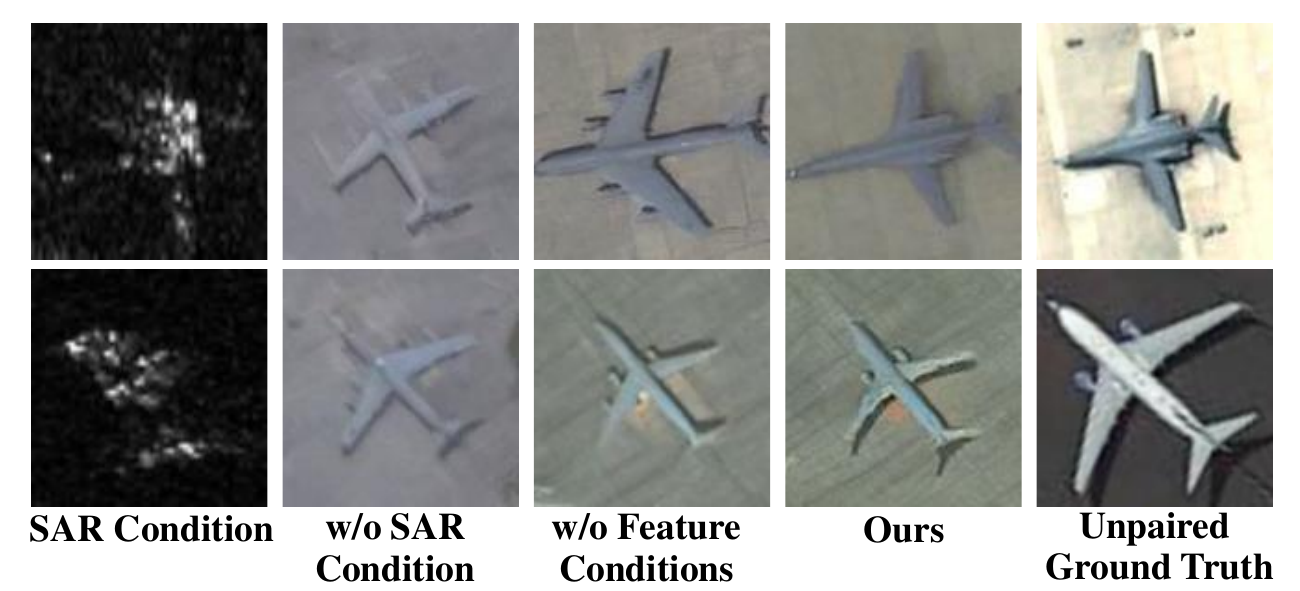}
\caption{Analysis of the impact of the input conditions on translation results.}

  \label{fig:no_sar}
\end{figure}

The denoising model relies on two crucial inputs: the SAR image and the feature condition provided by the detector.  Figure~\ref{fig:no_sar} illustrates the results of the experimental analysis of their individual contributions.

When the model is conditioned solely on the feature condition, including the detected category and keypoints, the generated images are heavily influenced by the quality of the detector’s predictions. If the detector is confident, the generated images accurately reflect the aircraft category and orientation. However, when the detector’s confidence is low or its predictions are incorrect, the resulting images exhibit blurry features, misaligned keypoints, and incorrect shape and orientations. 

Conversely, when only the SAR image is used without additional feature conditions, the model lacks the explicit guidance on target attributes, leading to inaccuracies in critical features as well. These include misidentifications of the aircraft type, wrong engine count, and inability to correctly model the target’s geometry including misaligned wing positions or body angles. Without feature information, the model generates SAR-to-OPT translations that are visually plausible in terms of texture and structure but fail to respect the semantics of the underlying target class.

These observations confirm the importance of both the SAR image condition and the feature condition for achieving accurate and realistic image generation. The interplay between these two sources of information ensures that the generated images not only resemble real optical images but also adhere to the correct target properties including category and angle.

\paragraph{Zero-Shot Generation Ability}  
% The proposed method demonstrates the capability to predict unseen aircraft categories based on the detector's keypoint heatmap matrix and category vector, enabling zero-shot generation. This is possible because the training set includes 15 aircraft categories with diverse engine counts, shapes, and orientations, which provide a broad representation of the target space. The detector's initial encoding effectively maps these features into the latent space, facilitating zero-shot generalization.

% Compared to methods like classifier-guided diffusion, which rely on explicit category supervision, the proposed approach sacrifices some accuracy for seen categories but gains significant generalization ability. This trade-off is valuable for extending the model's applicability to unseen categories. Figure~\ref{fig:zero_shot} showcases the results for two untrained aircraft types, Boeing 777 and Boeing 747, demonstrating that the generated images accurately capture basic attributes such as engine count, wing position, and orientation.
One of the key strengths of the proposed method is its ability to handle zero-shot generation, where the model can generate images for previously unseen categories. This capability arises from the ability to infer the category-specific attributes such as wing position, engine count and body shape, and align them within the latent space, even when no explicit training data for the new categories is available. The proposed model is trained on 15 distinct aircraft categories, each with a unique combination of aircraft attributes which together form a comprehensive representation of the target space.

During inference, when presented with a new category, the detector’s predictions of category vector and keypoint heatmap are used to guide the generation process. In comparison with methods  which require explicit category supervision for each generated image, the proposed method sacrifices some accuracy in the generation of seen categories to gain substantial generalization ability. This trade-off proves to be valuable, as it allows the model to handle a wider range of unseen aircraft categories, expanding its practical applicability. 

\begin{figure}[tb]
  \centering
\includegraphics[width=\linewidth]{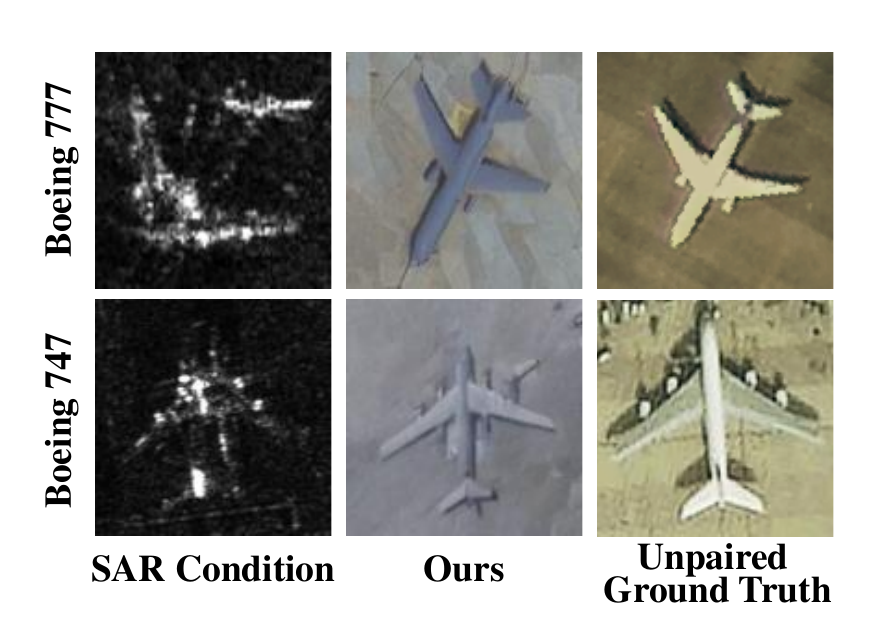}
\caption{Zero-shot generation results for  Boeing 777 and Boeing 747.}

  \label{fig:zero_shot}
\end{figure}

Figure~\ref{fig:zero_shot} presents the results of the zero-shot generation process, where the model successfully generates realistic optical images for aircraft types Boeing 777 and Boeing 747, which were not part of the training set. The generated images accurately reflect key attributes such as engine count, wing position, and body orientation, showcasing the model’s robustness and its capacity for generalization to  unseen categories. These findings suggest that the proposed method is not only effective for translating known categories but also capable of handling new categories in a zero-shot manner, making it a flexible and adaptable solution for real-world applications.

\section{Conclusion}
\label{sec:conclusion}

This study proposes a classifier-free guidance diffusion model for object-level SAR-to-optical image translation, addressing challenges including unpaired data and category-specific contour and texture restoration. A keypoints-guided conditional diffusion model is proposed, introducing a keypoints-based supervision framework that enables a pseudo-pairing training strategy and a keypoint-driven sampling process. Additionally, a Class-Angle Guidance Module (CAGM) is designed to encode target category and orientation information into the generation process. To ensure both the fidelity of image quality and the accuracy of target-specific features, consistency loss and adversarial loss are introduced, tailored to the requirements of the image translation task for aircraft.

Experimental results demonstrate the model's capability for object-level SAR-to-optical image translation of aircraft. Furthermore, the method's effectiveness is validated on untrained categories, showcasing its zero-shot generalization ability. Although the current model exhibits limitations in the precise translation of similar categories, future work will explore the potential of diffusion models in SAR-to-optical image translation, aiming to further enhance their performance. The application of such methods to downstream tasks will also be investigated, enabling more accurate fine-grained SAR image interpretation.

\bibliographystyle{IEEEtran}
\bibliography{references_sar2opt,references_unpairedI2I,references_diffusion,references_inrto}

\vfill
\end{document}